\documentclass{article}
\usepackage{acra}

\usepackage{amsmath,amssymb,amsfonts}
\usepackage{algpseudocode}
\usepackage{algorithm}		
\usepackage{graphicx}
\usepackage{textcomp}
\usepackage{xcolor}
\DeclareMathOperator*{\argmax}{argmax}

\begin{document}
\onecolumn
\noindent Pre-print of article that will appear in Proceedings of the Australasian Conference on Robotics and Automation 2018.

\vspace{\baselineskip}

\noindent Please cite this paper as:

\vspace{\baselineskip}

\noindent Stephen Hausler, Adam Jacobson, and Michael Milford. Feature Map Filtering: Improving Visual Place Recognition with Convolutional Calibration. Proceedings of Australasian Conference on Robotics and Automation, 2018.

\vspace{\baselineskip}

\noindent bibtex:

\vspace{\baselineskip}

\noindent @inproceedings\{hausler2018FeatFilt,\\
  \indent author = \{Hausler, Stephen and Jacobson, Adam and Milford, Michael\},\\
  \indent title = \{Feature Map Filtering: Improving Visual Place Recognition with Convolutional Calibration\},\\
  \indent booktitle = \{Proceedings of Australasian Conference on Robotics and Automation (ACRA)\},\\
  \indent year = \{2018\},\\
\}

\twocolumn

\title{Feature Map Filtering: Improving Visual Place Recognition with Convolutional Calibration}

\author{Stephen Hausler, Adam Jacobson and Michael Milford \\ Queensland University of Technology, Australia \\ stephen.hausler@hdr.qut.edu.au \thanks{SH is supported by a Research Training Program Stipend and ARC Future Fellowship FT140101229. AJ is supported by an Advance Queensland Innovation Partnership, Caterpillar and Mining3. MM is with the Australian Centre for Robotic Vision and was partially supported by an ARC Future Fellowship FT140101229.}
}

\maketitle

\begin{abstract}
Convolutional Neural Networks (CNNs) have recently been shown to excel at performing visual place recognition   under changing appearance and viewpoint. Previously, place recognition has been improved by intelligently selecting relevant spatial keypoints within a convolutional layer and also by selecting the optimal layer to use. Rather than extracting features out of a particular layer, or a particular set of spatial keypoints within a layer, we propose the extraction of features using a subset of the channel dimensionality within a layer. Each \emph{feature map} learns to encode a different set of weights that activate for different visual features within the set of training images. We propose a method of calibrating a CNN-based visual place recognition system, which selects the subset of feature maps that best encodes the visual features that are consistent between two different appearances of the same location. Using just 50 calibration images, all collected at the beginning of the current environment, we demonstrate a significant and consistent recognition improvement across multiple layers for two different neural networks. We evaluate our proposal on three datasets with different types of appearance changes - afternoon to morning, winter to summer and night to day. Additionally, the dimensionality reduction approach improves the computational processing speed of the recognition system. 
\end{abstract}

\begin{figure}[h!]
\includegraphics[scale=0.42,trim=0.4cm 1.0cm 0.2cm 0.1cm,clip]{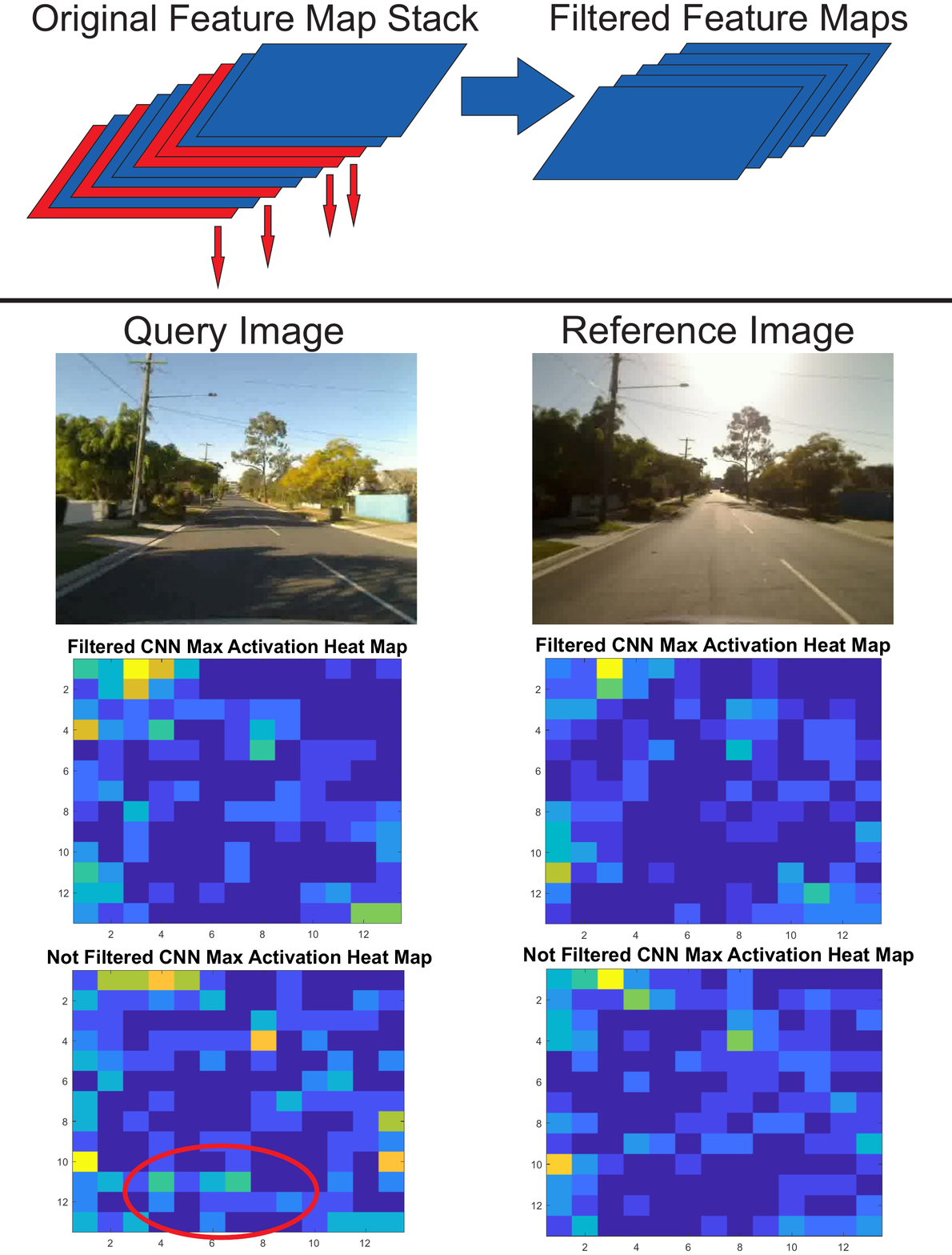}	
\caption{\emph{Top:} Our feature map filtering approach removes specific feature maps that vary their activation when the scenes' appearance changes over time. \emph{Bottom:} We display the maximum activation heat maps for the stack of feature maps that survive after filtering (the top row of heat maps), and for the removed feature maps (bottom row). Notice how the removed maps are activating in response to the shadows on the road (marked by a red ellipse).}
\label{TitlePage}
\end{figure}

\section{Introduction}

Visual Place Recognition, the ability to localize using just a visual sensor, is challenging due to the significant appearance change that visual scenes experience on a regular basis, including day to night, summer to winter and morning to afternoon. Both hand-crafted features, such as SURF \cite{BH2008} and HOG \cite{DN2005}, and deep learnt networks have been used to attempt to solve the VPR challenge \cite{NT2018,CM2008,SN2015,AR2018}. Both viewpoint and appearance robustness has been demonstrated when convolutional neural networks (CNNs) are used for visual place recognition \cite{SN2015}. This is especially the case when a CNN is trained for recognizing a specific environment \cite{AR2018,FeatReweight2017,ConvLearn2017}. However, this performance has the disadvantage of requiring training for all the environmental conditions that the robot is expected to experience, where-as for practical autonomy, the robot should be able to automatically, and swiftly, adjust its neural parameters to suit the current conditions.

We propose a novel solution to achieve this, by calibrating a convolutional neural network for the current environment. In state-of-the-art approaches, a neural network is re-trained for the specific environment by selecting a set of images from the new environment and re-training the model using these images \cite{AR2018,FeatReweight2017,ConvLearn2017}. However, this requires a significant time and processing cost, so much so that typical robot platforms do not have the capability to re-train the neural model online and in real-time. We propose a method that enables a fast, computationally cheap process of \emph{filtering} the collection of feature maps within a layer of a deep convolutional neural network (see Fig. \ref{TitlePage}). When a network is trained on a diverse set of images, each feature map encodes a different type of abstraction from this collection of images. For example, one map in a late convolutional layer might learn to `fire' upon regions of an image containing a building. We propose a calibration procedure which removes the feature maps that do not suit the recognition between the current environment and the learnt environment. This is achieved by minimizing the L2-distance between two identical locations that appear significantly different due to a change in the environment, while maximizing the distance between two different locations that look visually similar due to having the same environmental conditions.

We demonstrate the versatility of our approach by experimenting with two different CNN architectures, HybridNet \cite{CZ2017} and AlexNet trained on ImageNet \cite{AlexNet}, across three different datasets which demonstrate different types of appearance variations.

The paper proceeds as follows. In Section 2, we review prior uses of convolutional neural networks for the visual place recognition task and previous methods of neural network simplification. Section 3 presents our approach, explaining our calibration procedure and computational methods in detail. Section 4 details the setup of our experimental datasets and Section 5 evaluates the performance of feature map filtering, compared to not filtering. Section 6 provides intuition as to why feature map filtering works and Section 7 summarizes our contributions and provides suggestions for future work.

\section{Related Work}

In early experiments using convolutional neural networks for place recognition, a feature vector is produced from a particular layer of the network, using all the information that is encoded in the activations of that layer \cite{SN2015}. However, such a whole-image approach is sensitive to viewpoint variations. This was addressed by developing a landmark extraction algorithm and computing the neural responses to each landmark region in a scene \cite{Sunderhauf2015}. Intelligently selecting the useful information within an image is a valuable method of improving the localization performance. Rather than finding regions, LoST \cite{Lost2018} creates a feature vector by extracting semantically meaningful keypoints within the feature map spatial region. \cite{LookOnce2017} finds keypoints by observing the activations out of a late convolutional layer, while \cite{Zetao2018} trains a soft attention mask to select salient regions within an image to improve the selection of features used to formulate the feature vector. These keypoint feature vectors consist of the activations across all the feature maps within that layer at the spatial location of the keypoint, even if some of the feature maps are encoding visual information that is counter-productive to localizing in the current environment.

Several experiments compared the performance across different layers \cite{SN2015,CZ2017}, while a number of experiments use multiple layers simultaneously \cite{VehicleMultiLayer,Zetao2018}, to improve the visual recognition performance beyond the performance of a single layer. Different layers have been found to encode different types of visual features, such as color and texture in early layers, and objects and scenes in later layers \cite{NetDissect}.

The literature discussed in the previous paragraphs optimizes either the choice of layer to use, or the choice of spatial locations across the feature map stack. The third dimension to optimize is the choice of feature maps themselves within the stack of feature maps that comprise a layer. \cite{FeatPruneFirst2017} proposes that a CNN can be simplified by pruning the selection of feature maps, which attains comparable performance while improving the computational speed of a forward-pass through the network. \cite{FeatPrune2018} suggests an improvement by using linear discriminant analysis to calculate the discriminability score for each feature map. They are able to remove a greater number of feature maps without causing a major reduction in accuracy. \cite{ReThink} re-weights feature maps using a feedback process to improve the classification performance. However, they only re-weight feature maps and don't completely remove any feature maps. The concept of improving visual place recognition by discriminatively selecting a subset of the feature maps within a convolutional layer is a gap in the literature.

Recent literature on network dissection has provided evidence that individual feature maps encode specific visual features that are relatable to the classifier outputs \cite{NetDissect}. In their work, the hidden convolutional layers are probed by testing an individual feature map on a pixel-wise semantic segmentation task. They discover that individual feature maps activate for different objects, scenes, textures and colors. This research underpins the motivation for this work - for example, if a particular feature map activates to man-made lighting, this feature map will confuse the localization between night and day and is better removed from the feature vector.

\section{Proposed Approach}

\begin{figure}[t]
\centering
\includegraphics[width=\linewidth,trim=0cm 13.5cm 0cm 0.5cm,clip]{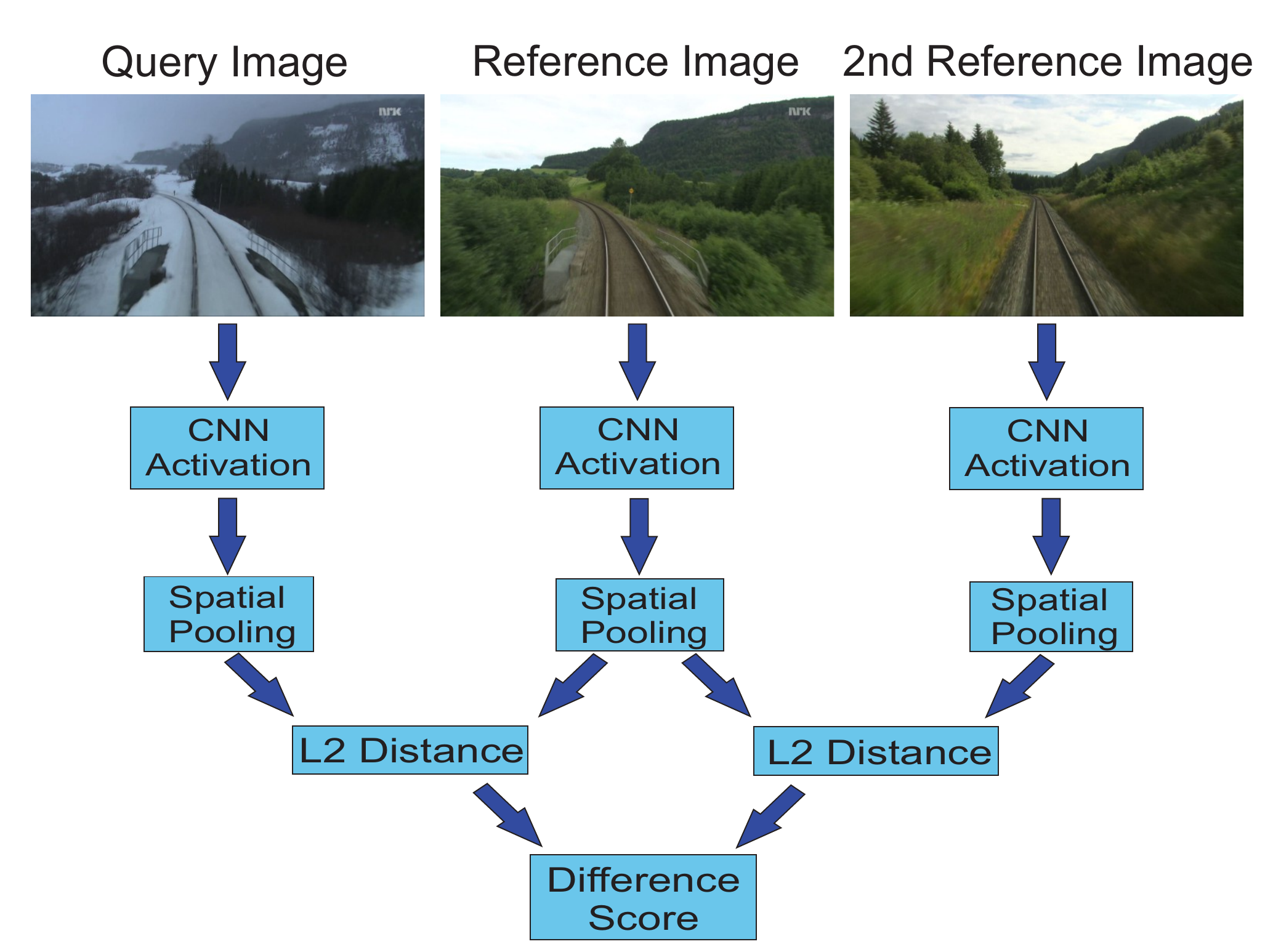}	
\caption{This diagram visually explains the triplet calibration method we employ.}
\label{MethodDiagram}
\end{figure}

We propose a novel method of calibrating a convolutional neural network for the current environment. Our calibration procedure removes the feature maps that do not suit the recognition between the current environment, and the learnt environment. This is achieved by minimizing the L2-distance between two identical locations that appear significantly different due to a change in the environment, while maximizing the distance between two different locations that look visually similar due to having the same environmental conditions (see Fig. \ref{MethodDiagram}). This is termed triplet loss in literature and like previous work, we also use the L2-distance as our calibration optimization metric \cite{FaceNetFirstTriplet,LightCNN,AR2018}. 

\subsection{Calibration Procedure}

For each calibration scene, we extract deep-learnt features for the currently viewed scene, the corresponding reference image and a randomly selected image elsewhere in the database of reference images. We use a total of 50 calibration images, all extracted from the beginning of the query dataset - this is to mimic the real world situation where the calibration is performed prior to the robot beginning the navigation of its environment. This calibration can be achieved using pre-defined maneuvers, such as the methods described in \cite{Jacobson2015}. These calibration triplets are used to perform feature map filtering, as explained in the following sections.

\subsection{Extracting Deep Learnt Features}

In a convolutional neural network, a convolutional layer tensor consists of dimension $W \times H \times C$, where $W$ and $H$ are the width and height of the data matrix and $C$ is the number of channels, otherwise termed the number of feature maps. To reduce the dimensionality of this $W \times H \times C$ feature vector, we use maximum pyramid spatial pooling \cite{CZ2017}, which was chosen as it has both viewpoint robustness and provides a significant dimensionality reduction while keeping the key features in each feature map. In our version of pyramid spatial pooling, we convert each $W \times H$ map into a vector of length 5, consisting of the maximum activation in each map and the maximum activation in each of the four quadrants of each $W \times H$ map. 

Out of a stack of feature maps within a convolutional layer, certain feature maps will activate to certain visual features in an image. For example, a feature map in a network might fire on regions of an image containing vehicles. However, in the context of visual place recognition, activations on vehicles has a negative effect, because vehicles are dynamic objects and not temporally static. This applies to other time-varying features, like snow in winter. Our goal is to search through the stack of feature maps to find the worst feature maps. We define the worst feature maps as feature maps that contain activations that vary across a change in appearance when the location does not change. We perform this search on the spatially pooled features in each feature map, for improved viewpoint robustness. 

\subsection{Filtering Feature Maps}

We use a Greedy algorithm \cite{Fegaras1998} to determine which subset of the feature map stack suits the current environmental conditions. Combinatorial optimization problems are typically NP-hard, with a variety of techniques employed to produce approximate solutions in related problems such as sensor selection \cite{SensorSelect2009}. In our method, using Greedy causes the worst feature map to be filtered at each iteration of the algorithm, until a local maximum is reached. We chose Greedy as it runs in polynomial time and was found to converge to a satisfactory position. 

To measure the feature map performance, we select each feature map individually and remove it from the feature vector before calculating the L2 (Euclidean) distance between both the images from the same location and the two images from the reference traverse. This results in two distance scores, one for the same location at different times of day and one for different locations at the same time of day. The result is a vector of difference scores across a different feature map being removed. 

\begin{equation}
	D(q_i^j,r_i^j) = \sqrt{ \sum^M_{k=1} (q_i^j(k) - r_i^j(k))^2}
\end{equation}
\noindent where $M$ is the dimension of the filtered query feature vector $q_i^j$.
\begin{equation}
    D(j) = D(r_i^j,n_i^j) - D(q_i^j,r_i^j)\quad \forall j
\end{equation}
\noindent where $r_i^j$ represents the current location filtered reference feature vector and $n_i^j$ represents the filtered feature vector from a random image somewhere else within the reference image database. $j$ denotes the index of the currently filtered feature map.

We then find the maximum distance:
\begin{equation}
    maxval = \underset{1 \leq j \leq N}{\max} D(j)
\end{equation}
\begin{equation}
    worstFmap = \underset{1 \leq j \leq N}{\argmax}\:\: D(j)
\end{equation}
\noindent where N is the number of remaining feature maps.

The index of the maximum distance represents the feature map to be removed to achieve the greatest L2 difference between the images from the same location and the images from different locations. With this chosen feature map, we modify the original feature vector and remove this worst performing feature map before repeating the above algorithm for this new, filtered, feature vector. 

We iterate in this fashion until a local maximum is reached, that is, the largest L2 difference between the images at same location and the images at different locations (with the images at the same location being closer in L2 space than the images at different locations). In our initial experiments we observed that the gradient towards the local maximum becomes very small prior to reaching the maximum and a significant number of feature maps are filtered out. As an alternate, less aggressive filtering algorithm, we added a gradient minimum cut-off threshold, which we set to 0.1. When removing the worst-performing feature map, if the difference between the previous iteration difference score and the current difference is less than 0.1, we stop the iteration and use the current set of remaining feature maps.

For improved robustness and to prevent outliers, we use multiple calibration images. The choice of filtered feature maps is stored for all images and after the calibration procedure is finished, the number of times a particular feature map is removed is summed across all 50 calibration images. We then find the set of feature maps that were least chosen to be filtered out, and the number of final feature maps is equal to the maximum number of remaining feature maps in the set of calibration images. This heuristic was chosen based on the principle that the choice of remaining feature maps needs to be able to encode all the features within all the calibration images, else minor variations in the current environment will cause key visual features to be missed. The filtering procedure is designed to only remove the feature maps that are irrelevant or damaging to the ability to match between the two appearances of the same location. 

\subsection{Place Recognition Validation Algorithm}

We developed a single-frame place recognition algorithm to evaluate the improvement gained by using feature map filtering. The features extracted from both the query images and the reference database only include the particular feature maps that were chosen by the feature map filter calibration algorithm. Each query image is compared to the reference database using the cosine distance metric to create a difference vector with length equal to the number of reference templates. We then normalise the difference vector to the range 0.001 to 0.999, where 0.001 denotes a poor match and 0.999 denotes the best matching template. We calculate the quality of the best matching template using a method originally proposed in SeqSLAM \cite{MM2012}, where the quality score is the ratio between the score at the best matching template and the next highest score outside a window around the best matching template. Precision and Recall scores are then calculated across a swept set of quality threshold values.

\section{Experimental Method}

\begin{figure}[h]
\centering
\includegraphics[width=\linewidth,trim=0cm 19.0cm 3.7cm 2.5cm,clip]{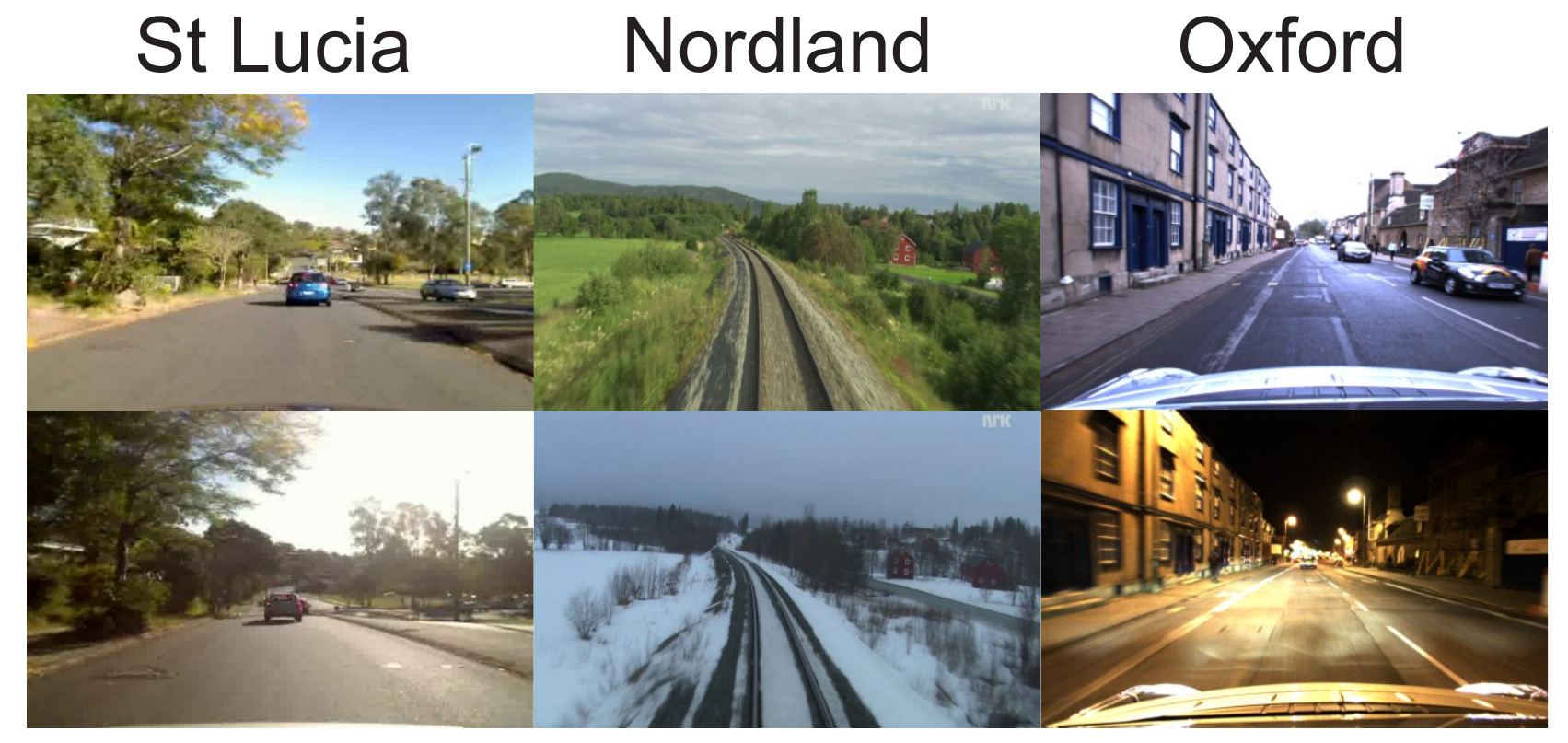}	
\caption{This panel of images displays example scenes from the St Lucia, Nordland and Oxford datasets at two different times. Notice the severe appearance change in all three examples.}
\label{DatasetExamples}
\end{figure}

We demonstrate our approach on three benchmark datasets, which have been extensively tested in recent literature \cite{SRAL,NT2018,GS2018}. Each dataset is briefly described in the sections below and visually shown in Figure \ref{DatasetExamples}.

\textbf{St Lucia} – consists of multiple vehicular traverses through the suburb of St Lucia, Brisbane across five different times of day \cite{LuciaDatasetRef}. We use the early morning traverse (190809\_0845) as the reference dataset and the late afternoon video (180809\_1545) as the query, with significant appearance change occurring between morning and afternoon. For the query traverse, we use 1000 images out of the original 15 FPS video. The dataset provides GPS ground truth and we use a ground-truth tolerance of 30 meters. For the calibration procedure, we extract 50 frames from the first 690 frames of the 15 FPS video. The query traverse is started after the last frame of the calibration procedure.

\textbf{Nordland} – The Nordland dataset \cite{NordlandDatasetRef} is recorded from a train travelling for 728 km through Norway across four different seasons. We use the Summer route as the reference dataset and the Winter traverse as the recognition route, using a 2000 image subset of the original videos. For the ground truth we compare the query traverse frame number to the matching database frame number, with a ground-truth tolerance of 10 frames, since the two traverses are aligned frame-by-frame. The 50 calibration images are collected from the videos immediately prior to the section we use for the 2000 image subset.

\textbf{Oxford RobotCar} - RobotCar was recorded over a year across different times of day, seasons and routes \cite{RobotCar}. We use an approximately 2 km route through Oxford, matching from an overcast day (2014-12-09-13-21-02) to night on the next day (2014-12-10-18-10-50). We down sample the original frame rate by a factor of three and start both traverses at the same location, corresponding to 1534 query images. We use a ground truth tolerance of 40 meters, consistent with a recent publication \cite{GS2018}. Calibration images are collected from the dataset prior to commencing the place recognition experiment.

\section{Results}

To produce our results, we run our filtering algorithm on layers Conv3 through to Conv6 of HybridNet and layers Conv2 through to Conv5 of AlexNet. By experimenting on multiple layers, the layer where filtering provides the greatest value can be found. The place recognition performance is evaluated using a single-frame matching algorithm and the F1 score metric is used to quantitatively measure the performance. In Tables \ref{HNetLuciaMapCounts} to \ref{ANetOxforMapCounts}, we compare the number of feature maps pre and post filtering and display the percentage filtered across different layers, networks and datasets.
\subsection{St Lucia}

\begin{figure}[h]
\centering
\includegraphics[width=\linewidth,trim=2cm 0cm 2cm 1.0cm,clip]{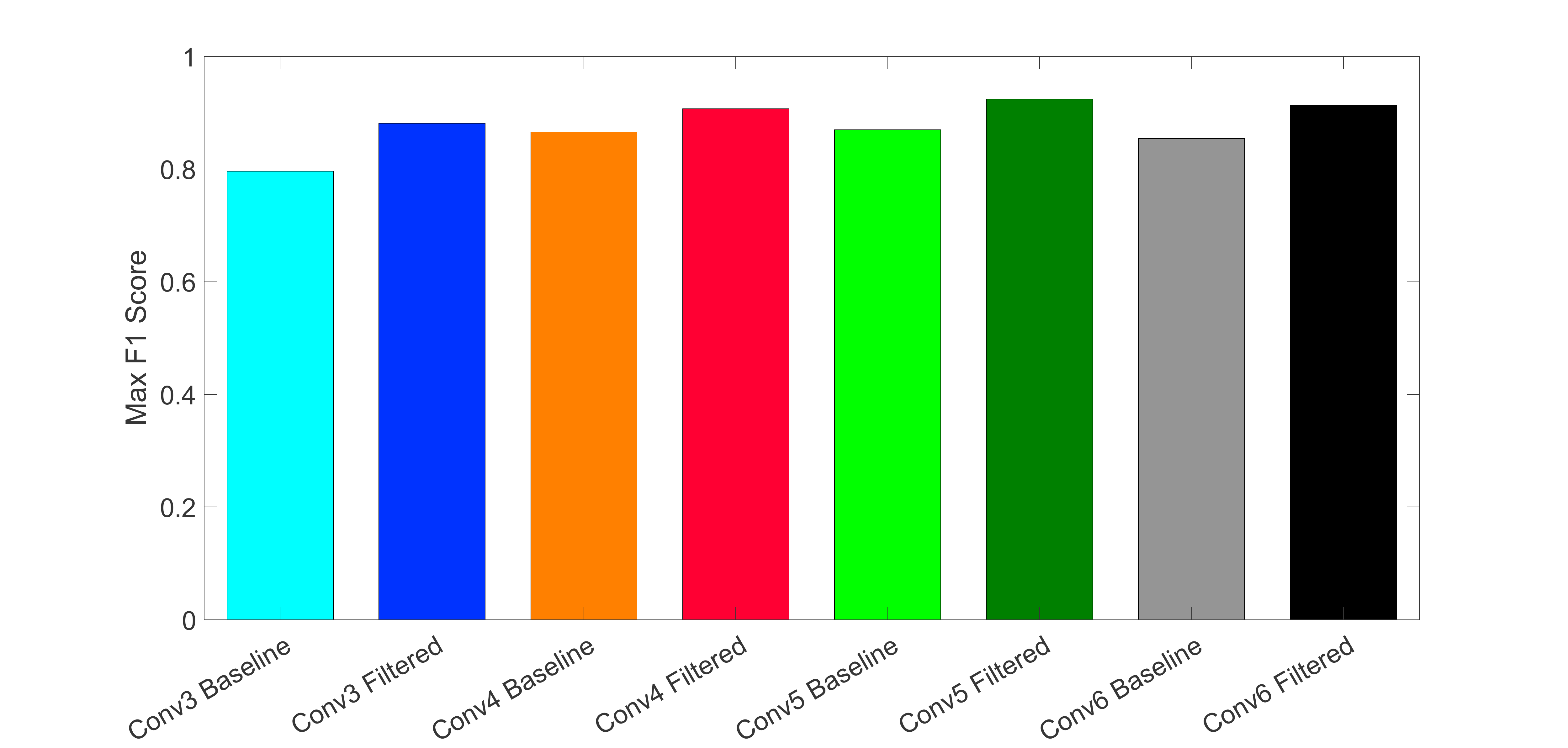}	
\caption{Maximum F1 score for Feature Map Filtering for HybridNet on the St Lucia dataset. We compare the filtered feature map recognition performance for four convolutional layers.}
\label{F1BarLuciaHNet}
\end{figure}

\begin{table}[h]
\caption{Number of feature maps pre-filtering and post-filtering for HybridNet on St Lucia}
\label{HNetLuciaMapCounts}
\centering
\begin{tabular}{|c|c|c|c|}
\hline\hline
\bfseries \small Layer & \bfseries \small Map Count &\bfseries \small Filtered Map Count & \bfseries \small \%\\
\hline
\small Conv-3 & 384 & 199 & 52\%\\
\hline
\small Conv-4 & 384 & 223 & 58\%\\
\hline
\small Conv-5 & 256 & 153 & 60\%\\
\hline
\small Conv-6 & 256 & 162 & 63\%\\
\hline
\end{tabular}
\end{table}

For HybridNet on St Lucia, filtering the stack of feature maps improves the localization performance across all layers (Fig. \ref{F1BarLuciaHNet}). This is to be expected when framed with respect to the original training data. HybridNet was trained on a collection of security cameras over time in disparate locations \cite{CZ2017}, thus certain feature maps would have learnt to encode visual features that enable matching between summer and winter while others learn to match from morning to afternoon. Since the class output of HybridNet classifies images to a particular location, this encoding is consistent even at higher network layers.

\begin{figure}[h]
\centering
\includegraphics[width=\linewidth,trim=2cm 0cm 2cm 1.0cm,clip]{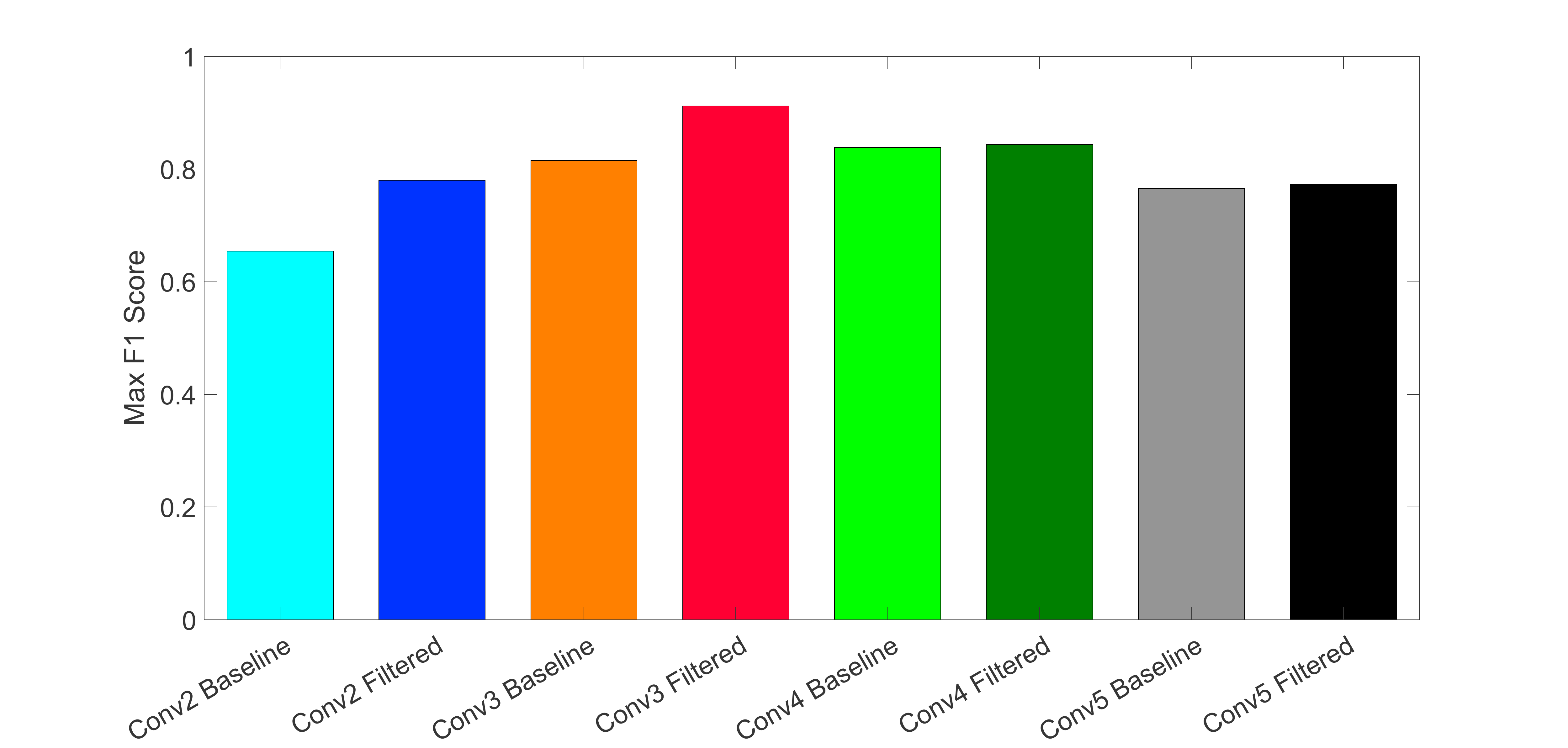}	
\caption{Maximum F1 score for Feature Map Filtering for AlexNet on the St Lucia dataset. We compare the filtered feature map recognition performance for four convolutional layers.}
\label{F1BarLuciaANet}
\end{figure}

When filtering is applied to AlexNet, unlike HybridNet, not all layers find a major improvement after filtering. Only Conv2 and Conv3 find a significant improvement using filtering (Fig. \ref{F1BarLuciaANet}). Also, a larger number of feature maps are filtered for the same gradient cut-off threshold. Since AlexNet is trained on a wider variety of images that are not applicable to visual place recognition (such as images of clothing), a larger proportion of feature maps need to be removed in the higher network layers (refer to Tables \ref{HNetLuciaMapCounts} and \ref{ANetLuciaMapCounts}).

\begin{table}[h]
\caption{Number of feature maps pre-filtering and post-filtering for AlexNet on St Lucia}
\label{ANetLuciaMapCounts}
\centering
\begin{tabular}{|c|c|c|c|}
\hline\hline
\bfseries \small Layer & \bfseries \small Map Count &\bfseries \small Filtered Map Count & \bfseries \small \%\\
\hline
\small Conv-2 & 256 & 129 & 50\%\\
\hline
\small Conv-3 & 384 & 174 & 45\%\\
\hline
\small Conv-4 & 384 & 169 & 44\%\\
\hline
\small Conv-5 & 256 & 113 & 44\%\\
\hline
\end{tabular}
\end{table}
\subsection{Nordland}

\begin{figure}[h]
\centering
\includegraphics[width=\linewidth,trim=2cm 0cm 2cm 1.0cm,clip]{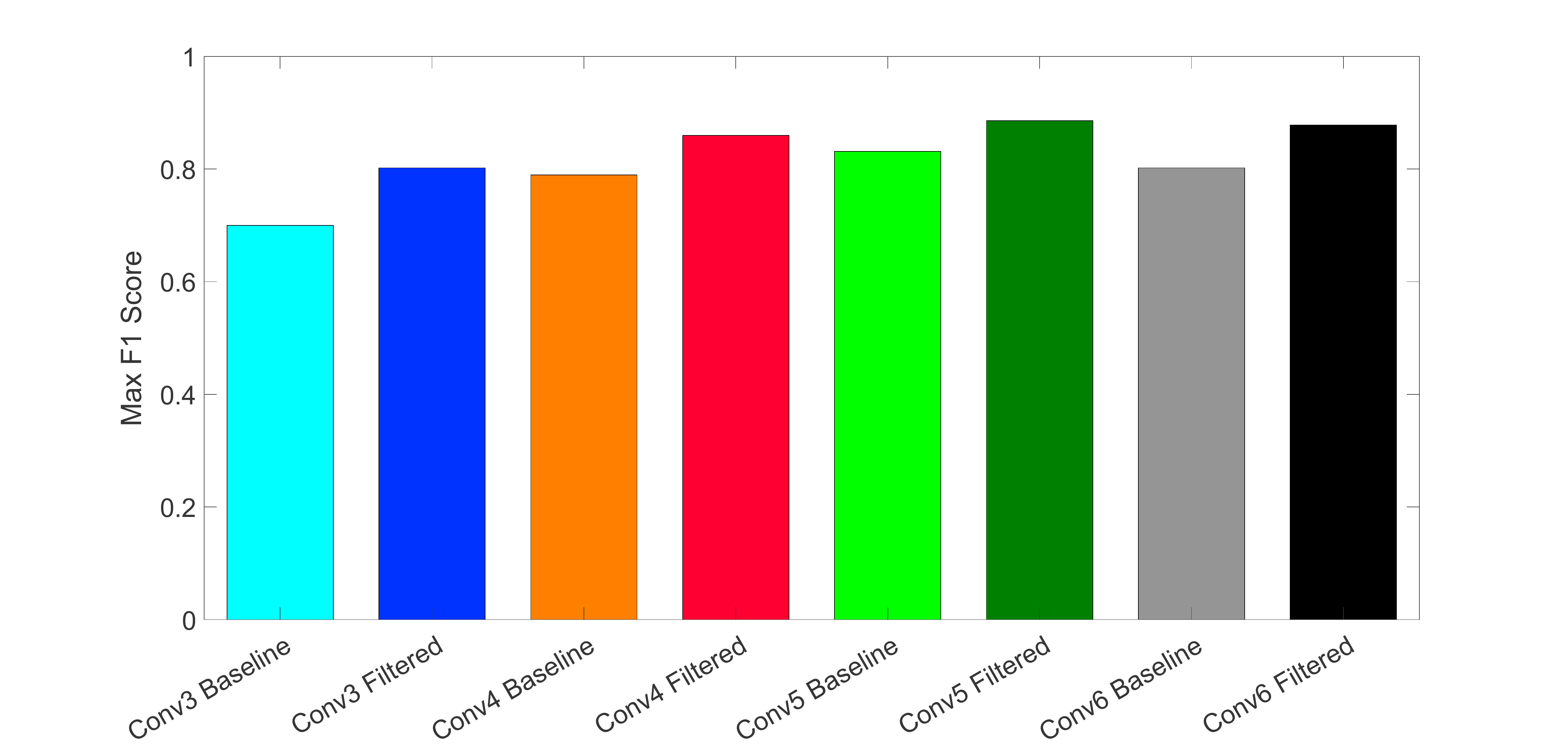}	
\caption{Maximum F1 score for Feature Map Filtering for HybridNet on the Nordland dataset. We compare the filtered feature map recognition performance for four convolutional layers.}
\label{F1BarNordHNet}
\end{figure}

Like our experiment on the St Lucia dataset, when filtering is applied to HybridNet, our recognition performance improves consistently across all four layers (Fig. \ref{F1BarNordHNet}). In this dataset, which has a greater appearance change, a larger number of feature maps are filtered for all four layers (see Table \ref{HNetNordlandMapCounts}). From the results we can also infer that the higher network layers are more appearance invariant, since proportionally less feature maps require filtering.

\begin{table}[h]
\caption{Number of feature maps pre-filtering and post-filtering for HybridNet on Nordland}
\label{HNetNordlandMapCounts}
\centering
\begin{tabular}{|c|c|c|c|}
\hline\hline
\bfseries Layer & \bfseries \small Map Count &\bfseries \small Filtered Map Count & \bfseries \small \%\\
\hline
\small Conv-3 & 384 & 150 & 39\%\\
\hline
\small Conv-4 & 384 & 167 & 43\%\\
\hline
\small Conv-5 & 256 & 125 & 49\%\\
\hline
\small Conv-6 & 256 & 150 & 59\%\\
\hline
\end{tabular}
\end{table}

\begin{table}[h!]
\caption{Number of feature maps pre-filtering and post-filtering for AlexNet on Nordland}
\label{ANetNordlandMapCounts}
\centering
\begin{tabular}{|c|c|c|c|}
\hline\hline
\bfseries Layer & \bfseries \small Map Count &\bfseries \small Filtered Map Count &\bfseries \small \%\\
\hline
\small Conv-2 & 256 & 90 & 35\%\\
\hline
\small Conv-3 & 384 & 132 & 34\%\\
\hline
\small Conv-4 & 384 & 160 & 42\%\\
\hline
\small Conv-5 & 256 & 111 & 43\%\\
\hline
\end{tabular}
\end{table}

\begin{figure}[h!]
\centering
\includegraphics[width=\linewidth,trim=2cm 0cm 2cm 1.0cm,clip]{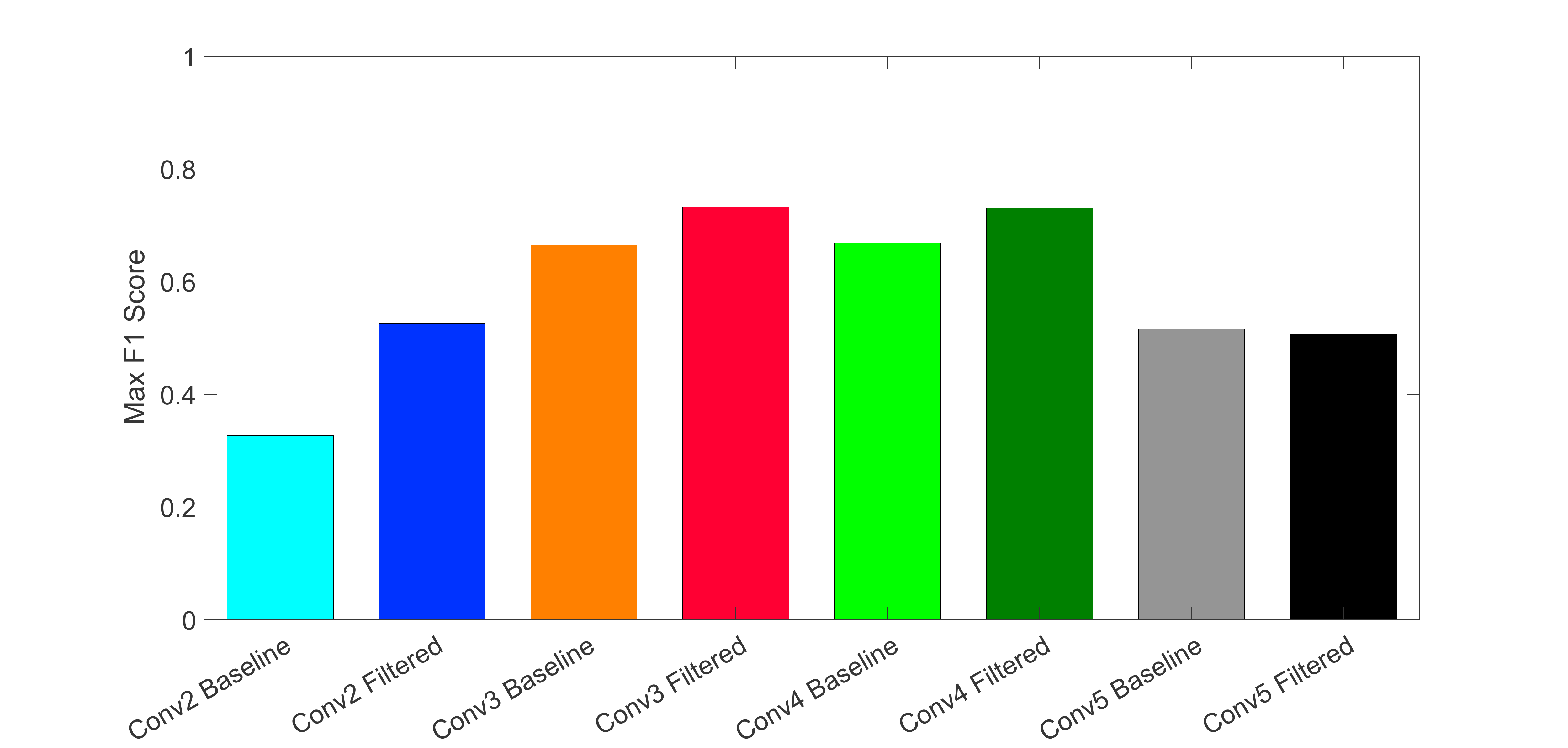}	
\caption{Maximum F1 score for Feature Map Filtering for AlexNet on the Nordland dataset. We compare the filtered feature map recognition performance for three convolutional layers.}
\label{ANetF1BarNord}
\end{figure}

When AlexNet is applied to the Nordland dataset, a larger proportion of feature maps require filtering (see Table \ref{ANetNordlandMapCounts}). As can be seen in Figure \ref{ANetF1BarNord}, for Conv2, Conv3 and Conv4, feature map filtering improves the baseline place recognition performance. The improvement is particularly apparent for Conv2. In related works \cite{SN2015,Sunderhauf2015,CZ2017}, Conv2 is not considered for place recognition and our baseline results reflect the typically poor performance using Conv2. However, when filtering is used, the place recognition performance exceeds that of Conv5 baseline.

\subsection{Oxford RobotCar}

\begin{figure}[h]
\centering
\includegraphics[width=\linewidth,trim=2cm 0cm 2cm 1.0cm,clip]{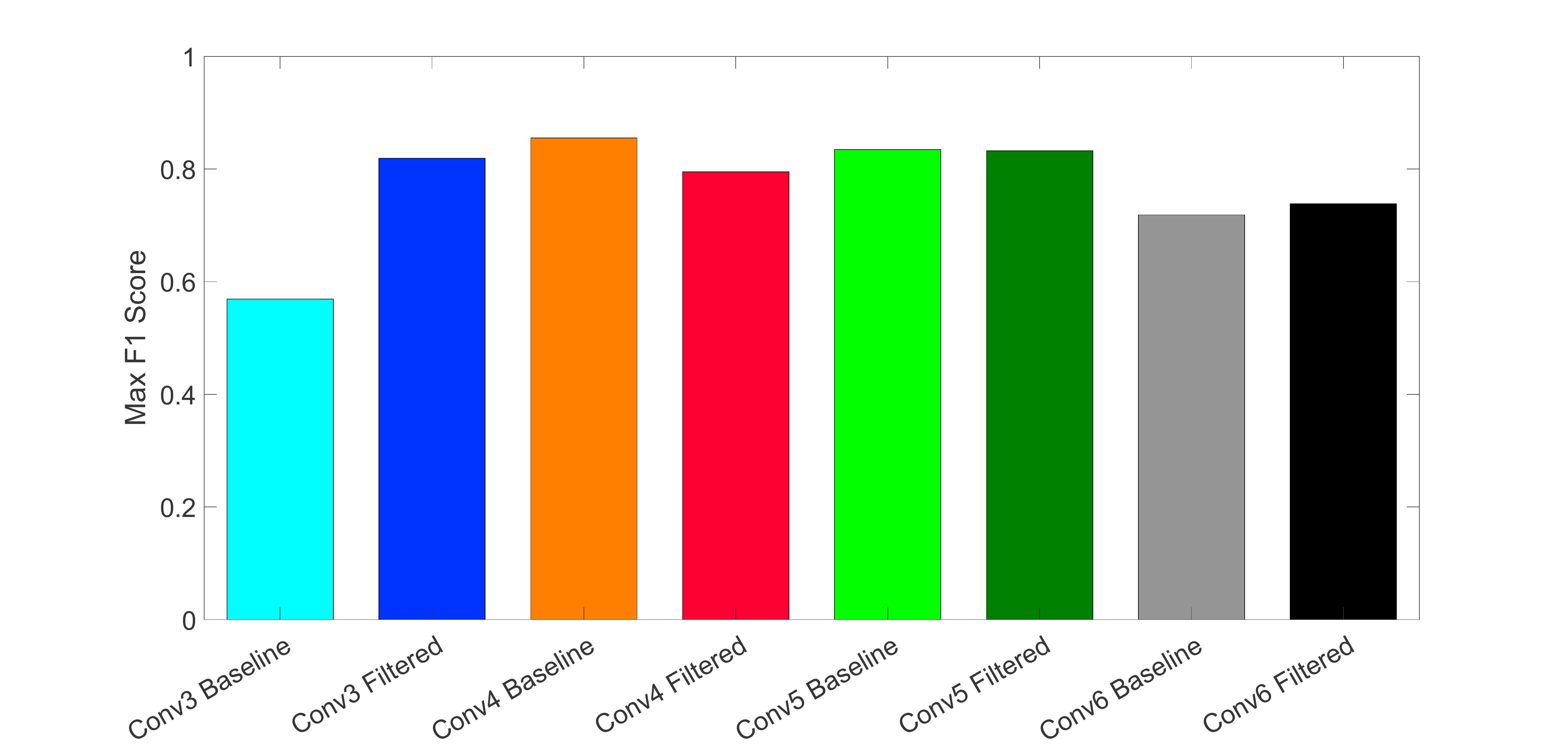}	
\caption{Maximum F1 score for Feature Map Filtering for HybridNet on the Oxford RobotCar dataset. We compare the filtered feature map recognition performance for four convolutional layers.}
\label{F1BarOxfordHNet}
\end{figure}

It is worth noting that for the same gradient cut-off threshold, more feature maps are filtered on the Oxford RobotCar dataset (see Table \ref{HNetOxforMapCounts}). We hypothesize that this is because this dataset has the greatest appearance variation of night to day. Filtering only improves Conv3 by a noticeable margin on the Oxford dataset (refer to Fig. \ref{F1BarOxfordHNet}). A possible explanation for this is a mismatch between the scene categories observed in the calibration images and the scenes observed in other sections of the dataset. For example, the calibration route occurs through an urban street with no vegetation, while later in the dataset, the road travels past a park. Conv3 encodes more generic visual features which are captured during the calibration route.

\begin{table}[h]
\caption{Number of feature maps pre-filtering and post-filtering for HybridNet on Oxford RobotCar}
\label{HNetOxforMapCounts}
\centering
\begin{tabular}{|c|c|c|c|}
\hline\hline
\bfseries Layer & \bfseries \small Map Count &\bfseries \small Filtered Map Count & \bfseries \small \%\\
\hline
\small Conv-3 & 384 & 117 & 30\%\\
\hline
\small Conv-4 & 384 & 137 & 36\%\\
\hline
\small Conv-5 & 256 & 112 & 44\%\\
\hline
\small Conv-6 & 256 & 134 & 52\%\\
\hline
\end{tabular}
\end{table}

\begin{figure}[h]
\centering
\includegraphics[width=\linewidth,trim=2cm 0cm 2cm 1.0cm,clip]{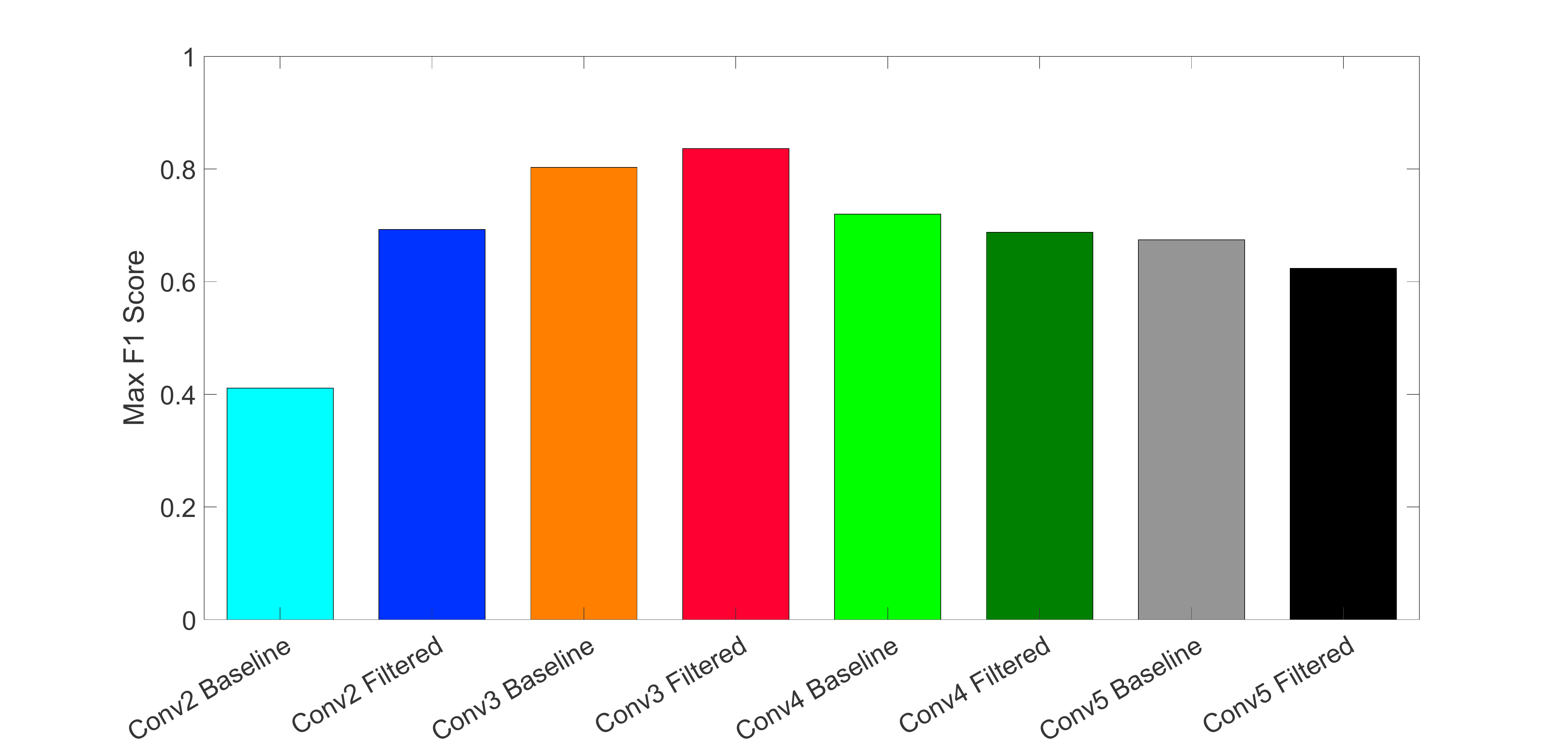}	
\caption{Maximum F1 score for Feature Map Filtering for AlexNet on the Oxford RobotCar dataset. We compare the filtered feature map recognition performance for four convolutional layers.}
\label{F1BarOxfordANet}
\end{figure}

\begin{table}[h]
\caption{Number of feature maps pre-filtering and post-filtering for AlexNet on Oxford RobotCar}
\label{ANetOxforMapCounts}
\centering
\begin{tabular}{|c|c|c|c|}
\hline\hline
\bfseries Layer & \bfseries \small Map Count &\bfseries \small Filtered Map Count & \bfseries \small \%\\
\hline
\small Conv-2 & 256 & 62 & 24\%\\
\hline
\small Conv-3 & 384 & 142 & 37\%\\
\hline
\small Conv-4 & 384 & 138 & 36\%\\
\hline
\small Conv-5 & 256 & 88 & 34\%\\
\hline
\end{tabular}
\end{table}

When our calibration procedure is applied to AlexNet, the same trend continues - the larger appearance variation causes a greater proportion of feature maps to be filtered out (refer to Table \ref{ANetOxforMapCounts}). For the Conv2 layer, three quarters of the original stack of feature maps are removed and in doing so, the maximum F1 score increases from 0.41 to 0.69. This is further evidence that our proposed approach is successfully finding the feature maps that are consistent across the appearance change. Again, the higher level layers gain no localization benefit from feature map filtering, however an improvement is still made to the compute time. 

\section{Discussion}

\subsection{Localization Improvement}

\begin{figure}[h]
\centering
\includegraphics[width=\linewidth,trim=0cm 0cm 0.5cm 0.0cm,clip]{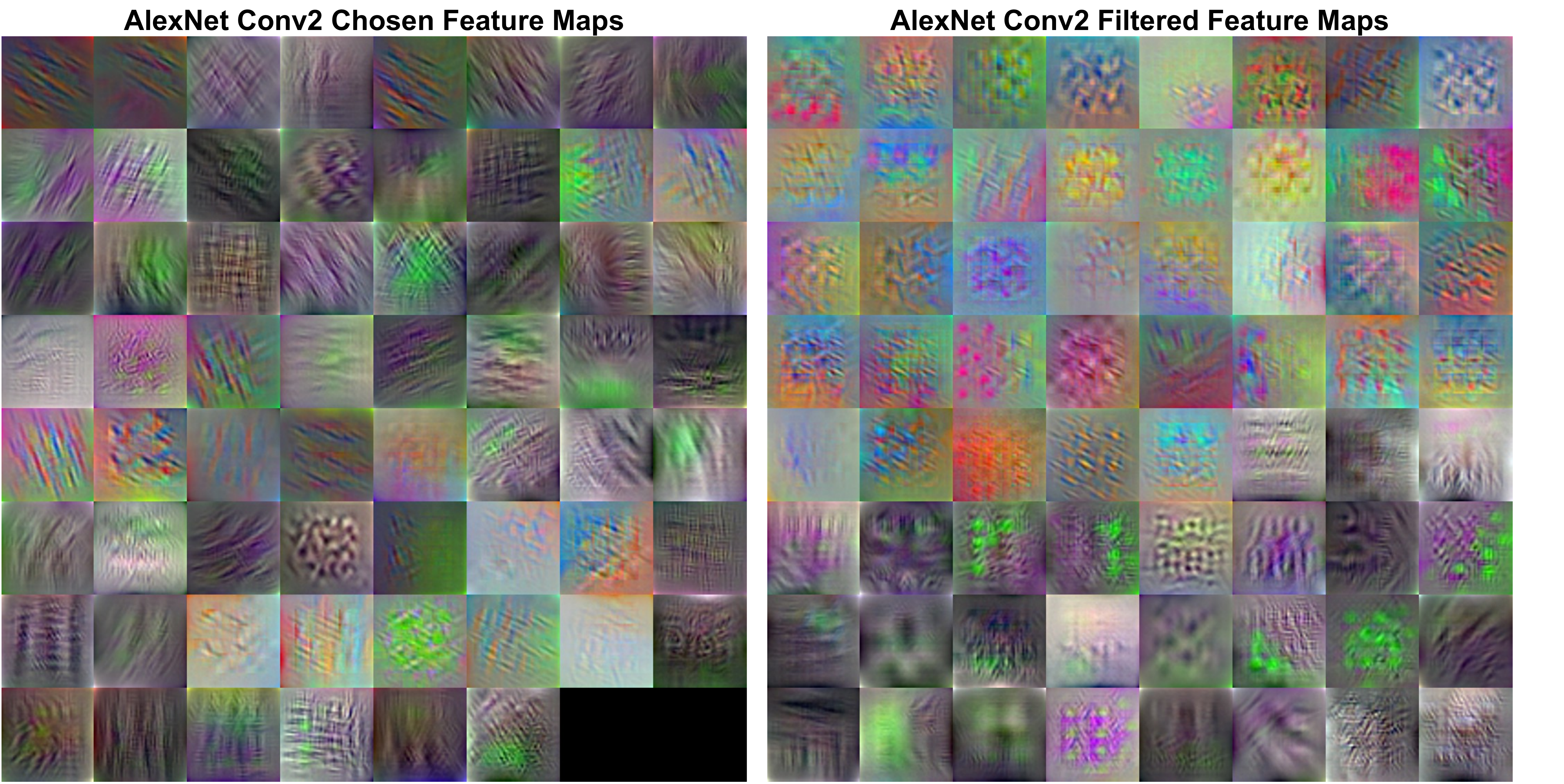}	
\caption{On the Oxford dataset and using Conv2 of AlexNet, 62 feature maps are selected post-filtering. Using MATLAB's deepDreamImage function, we visualize the types of visual features that the chosen feature maps respond to (left-hand montage) and compare against a selection of filtered feature maps (right-hand montage). Notice the similarity between the images in the left-hand montage and the presence of many `line-based' filters. This is explainable considering that street lighting is designed to illuminate road markings and road markings are typically straight line segments.}
\label{DeepDream}
\end{figure}

\begin{figure}[h]
\centering
\includegraphics[width=\linewidth,trim=0.5cm 4.4cm 0.4cm 1.2cm,clip]{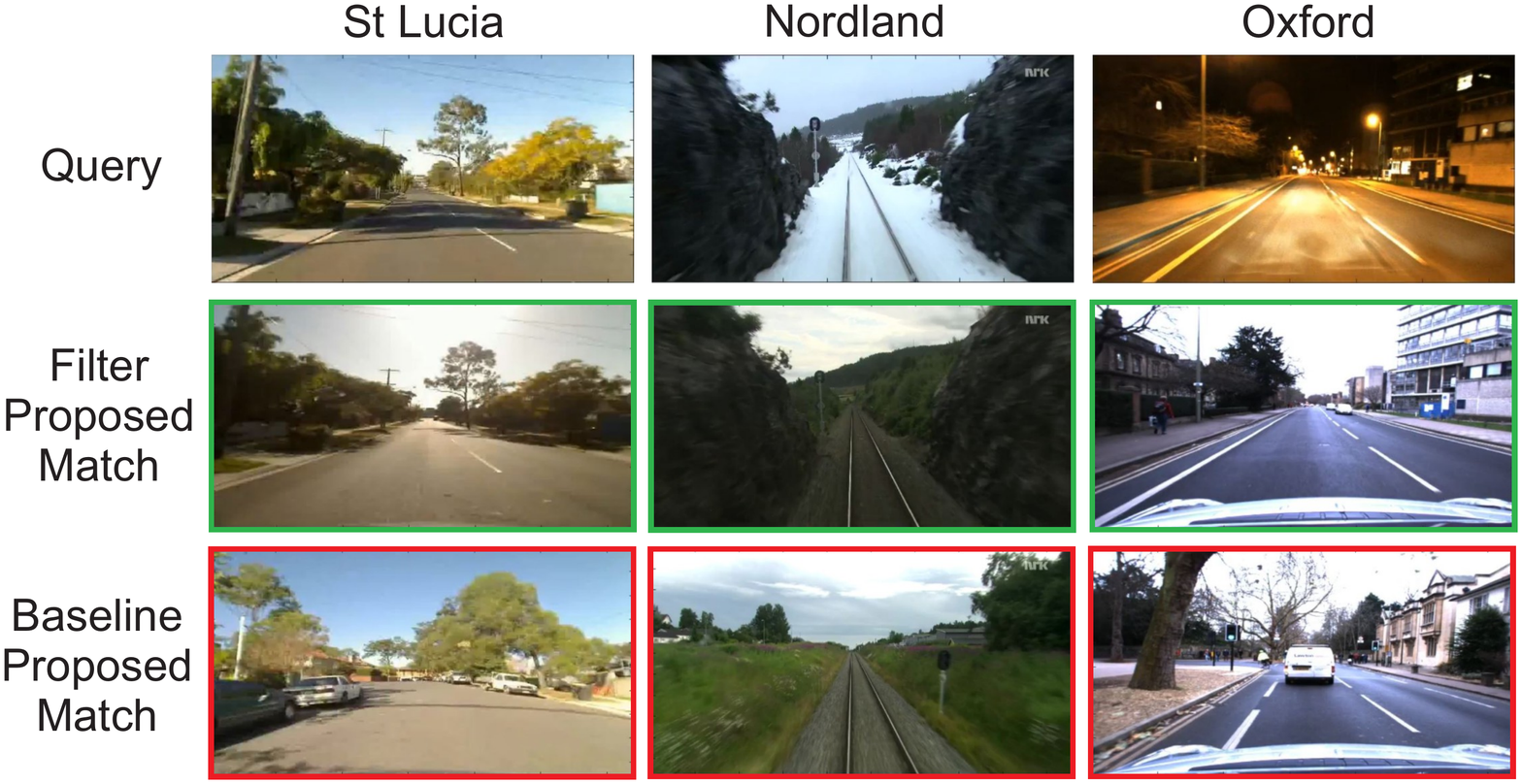}	
\caption{Examples on St Lucia, Nordland and Oxford where filtering the stack of feature maps enables successfull localization. The baseline match was generated using all the feature maps in Conv3 of HybridNet and the filter match is the correct location hypothesis when the filter calibration procedure is applied.}
\label{Recognition_Panel}
\end{figure}

A key result from our experimentation is that filtering provides a considerable improvement to earlier convolutional layers. Early layers have been shown to encode simple visual features while later layers encode objects and regions that are associated with the final class outputs \cite{NetDissect}. Our results show that filtering object types has less of an advantage, since objects within a scene are typically less affected by environmental changes than lower level visual features, such as the color of the leaves of a tree. When an early layer is filtered, filters that encode a visual feature that is impacted by the change in environment is removed, leaving only the visual features that remain consistent over time. The feature maps selected by our approach can be visually seen in Figure \ref{DeepDream}. We also show examples where our filtering approach enables localization when the baseline of not filtering causes an incorrect place hypothesis (see Fig. \ref{Recognition_Panel}).

\subsection{Computational Improvement}

Our improved F1 scores across most layers on both HybridNet and AlexNet is particularly significant when compared to the quantity of feature maps that are removed. As can be seen in the six tables, our filter algorithm filters, on average, 51\% of all feature maps when HybridNet is used and 61\% when AlexNet is used. This is a significant reduction of information and yet we achieve improved localization performance and significantly improve the place recognition computation time. For example, using Conv3 of HybridNet requires an average of 68 ms to match a query image to a reference database of 1442 images (on a standard desktop PC). When filtering is used, this drops to 43.9ms, 64\% of the original time per frame. This is even more apparent with Conv2 of AlexNet on the Oxford RobotCar dataset, where the processing time halves from 81ms to 41ms.

\section{Conclusion and Future Work}

This paper proposes a novel method of performing convolutional network calibration for visual place recognition, without requiring any computationally intensive re-training of the neural network parameters. We achieve this by filtering the set of feature maps produced by a layer within a CNN, by minimizing the L2 distance between the current scene and the corresponding reference image while maximizing the distance between the reference image and another reference image elsewhere in the database. Our feature map filtering approach has two key advantages: improved localization ability in changing environments, and improved computation speed. Our results demonstrate a considerable localization improvement for earlier network layers, with the greatest improvement on the Oxford RobotCar dataset, matching from night to day, using the Conv3 layer on HybridNet and the Conv2 layer on AlexNet. Our calibration procedure resulted in an improvement in HybridNet's Conv3 F1 score from 0.56 to 0.81 and AlexNet's Conv2 F1 score from 0.41 to 0.69.

Future work will devise a method of performing feature map filtering in real-time, without requiring any prior calibration. This could be achieved by devising a method of classifying the type of visual feature a particular feature map is activating to and specifically filtering the set of classes that are only occurring in the query traverse and not present anywhere in the reference traverse (such as street lighting at night-time). Also, our feature map calibration strategy using Greedy could be replaced with an alternative heuristic, to further improve the optimization quality. Finally, feature map filtering may also have applications in other computer vision tasks, as this approach could be used to quickly prepare a deep, generically trained CNN for a very specific task without re-training the network weights.

\bibliographystyle{named}
\bibliography{FeatFilt}

\end{document}